\documentclass{article}

\usepackage{microtype}
\usepackage{graphicx}
\usepackage{booktabs} % for professional tables

\usepackage{hyperref}
\usepackage[toc,page]{appendix}

\usepackage[accepted]{icml2020}
\usepackage{booktabs}
\usepackage{array}
\graphicspath{ {./images/} }
\newcolumntype{L}{>{\centering\arraybackslash}m{1.7cm}}
\usepackage{xcolor}
\usepackage{upgreek}
\usepackage{amsmath}
\usepackage{amssymb}
\usepackage{subcaption}
\usepackage{multirow}
\usepackage[utf8]{inputenc}
\usepackage{amsthm}
\usepackage{comment}
\usepackage{xspace}

\theoremstyle{definition}

\DeclareMathOperator{\E}{\mathbb{E}}

\newcommand{\todopb}[1]{}
\newcommand{\ipdistill}{$GameDistill$\xspace}

\icmltitlerunning{Inducing Cooperation in Sequential Social dilemmas through MARL using Status-Quo Loss}

\begin{document}

\twocolumn[
% \icmltitle{Inducing Cooperation in Multi-Agent Games Through Status-Quo Loss}
\icmltitle{Inducing Cooperative behaviour in Sequential-Social dilemmas through Multi-Agent Reinforcement Learning using Status-Quo Loss}




% \author{$^1$, $^1$, $^1$, $^2$*,\\ $^1$, $^1$}
% % $^1$IIIT-Hyderabad, India \\
% % $^2$Adobe Research, India}
% \affiliation{$^1$}
% \affiliation{$^2$}

\icmlsetsymbol{equal}{*}

\begin{icmlauthorlist}
\icmlauthor{Pinkesh Badjatiya}{adb}
\icmlauthor{Mausoom Sarkar}{adb}
\icmlauthor{Abhishek Sinha}{adb}
\icmlauthor{Siddharth Singh}{kgp}
\icmlauthor{Nikaash Puri}{adb}
\icmlauthor{Jayakumar Subramanian}{adb}
\icmlauthor{Balaji Krishnamurthy}{adb}
\end{icmlauthorlist}

\icmlaffiliation{adb}{Media and Data Science Research Lab, Adobe}
\icmlaffiliation{kgp}{IIT Kharagpur, India}

\icmlcorrespondingauthor{Pinkesh Badjatiya}{pbadjati@adobe.com}

\icmlkeywords{Reinforcement Learning, multi-agents, emergent cooperation}

\vskip 0.3in
]



\printAffiliationsAndNotice{}  % leave blank if no need to mention equal contribution

\begin{abstract}
    In social dilemma situations, individual rationality leads to sub-optimal group outcomes.
    Several human engagements can be modeled as a sequential (multi-step) social dilemmas.
    However, in contrast to humans, Deep Reinforcement Learning agents trained to optimize individual rewards in sequential social dilemmas converge to selfish, mutually harmful behavior.
    We introduce a status-quo loss ($SQLoss$) that encourages an agent to stick to the status-quo, rather than repeatedly changing its policy.
    We show how agents trained with $SQLoss$ evolve cooperative behavior in several social dilemma matrix games.
    To work with social dilemma games that have visual input, we propose $GameDistill$.
    $GameDistill$ uses self-supervision and clustering to automatically extract cooperative and selfish policies from a social dilemma game.
    We combine $GameDistill$ and $SQLoss$ to show how agents evolve socially desirable cooperative behavior in the Coin Game.
\end{abstract}

\section{Introduction}
\label{sec:introduction}

Consider a sequential social dilemma, where individually rational behavior leads to outcomes that are sub-optimal for each individual in the group. \cite{Hardin1968,ostrom:1990,Ostrom278,Dietz1907}. 
Current state-of-the-art Multi-Agent Deep Reinforcement Learning (MARL) methods that train agents independently can lead to agents that fail to cooperate reliably, even in simple social dilemma settings. 
This failure to cooperate results in sub-optimal individual and group outcomes (\citet{foerster2018learning, Peysakhovich1707.01068}, Section~\ref{sec:approach:selfish-learner}). 

\begin{figure}
    \centering
    \includegraphics[width=0.75\linewidth]{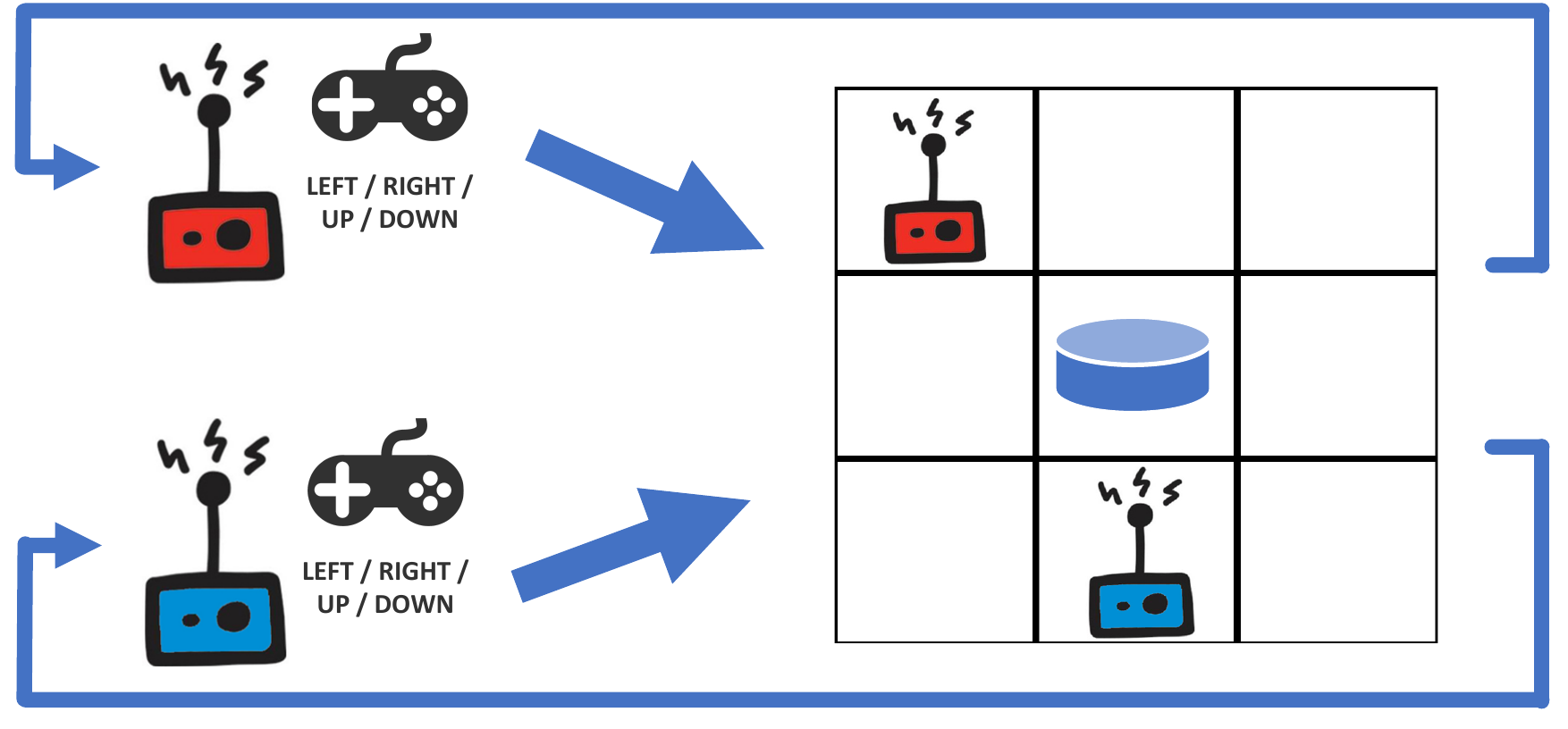}
    \caption{
    Two agents (Red and Blue) playing the Coin Game. 
    The agents, along with a Blue or Red coin, appear at random positions in a 3x3 grid. 
    An agent observes the complete 3x3 grid as input and can move either left, right, up, or down.
    When an agent moves into a cell with a coin, it picks the coin, and a new instance of the game begins.
    If the Red agent picks the Red coin, it gets a reward of $+1$ and the Blue agent gets no reward.
    If the Red agent picks the Blue coin, it gets a reward of $+1$, and the Blue agent gets a reward of $-2$.
    The blue agent's reward structure is symmetric to that of the red agent.
    }
    \label{fig:coin_game}
\end{figure}

To illustrate why it is challenging to evolve cooperation in such dilemmas, we consider the Coin Game (\cite{foerster2018learning}, Figure~\ref{fig:coin_game}). 
Each agent can play either selfishly (pick all coins) or cooperatively (pick only coins of its color). 
Regardless of the behavior of the other agent, the individually rational choice for an agent is to play selfishly, either to minimize losses (avoid being exploited) or to maximize gains (exploit the other agent).  
However, when both agents behave rationally, they try to pick all coins and achieve an average long term reward of $-0.5$. 
In contrast, if both play cooperatively, then the average long term reward for each agent is $0.5$.
Therefore, when agents cooperate, they are both better off. 
Training Deep RL agents independently in the Coin Game using state-of-the-art methods leads to mutually harmful selfish behavior (Section~\ref{sec:approach:selfish-learner}).

In this paper, we present a novel MARL algorithm that allows independently learning Deep RL agents to converge to \textbf{individually and socially desirable cooperative behavior} in such social dilemma situations. 
Our key contributions can be summarised as:
\begin{enumerate}
    \item We introduce a \textbf{Status-Quo} loss ($SQLoss$, Section~\ref{sec:approach:SQLoss}) and an associated policy gradient-based algorithm to evolve optimal behavior for agents that can act in either a cooperative or a selfish manner, by choosing between a cooperative and a selfish policy. 
     We empirically demonstrate that agents trained with the $SQLoss$ evolve optimal behavior in several social dilemma iterated matrix games (Section \ref{sec:results}).
    \item We propose \ipdistill (Section~\ref{sec:GameDistill}), an algorithm that reduces a social dilemma game with visual observations to an iterated matrix game by extracting policies that implement cooperative and selfish behavior.  
    We empirically demonstrate that \ipdistill extracts cooperative and selfish policies for the Coin Game (Section~\ref{sec:results:visual-input-gamedistll-and-sqloss}). 
    \item We demonstrate that when agents run \ipdistill followed by MARL game-play using $SQLoss$, they converge to individually as well as socially desirable cooperative behavior in a social dilemma game with visual observations (Section~\ref{sec:results:visual-input-gamedistll-and-sqloss}).
\end{enumerate}

The problem of how independently learning agents evolve cooperative behavior in social dilemmas has been studied by researchers through human studies and simulation models~\cite{Fudenberg, Edward:1984, Fudenberg:1984, Kamada:2010, Abreu:1990}. 
A large body of work has looked at the mechanism of evolution of cooperation through reciprocal behaviour and indirect reciprocity~\cite{reciprocal1971,reciprocal1984,reciprocal1992,reciprocal1993,in_reciprocal1998}, through variants of reinforcement using aspiration~\cite{reinforce_variant}, attitude~\cite{NonRL_attitude} or multi-agent reinforcement learning~\cite{Sandholm1996MultiagentRL,Wunder:2010}, and under specific conditions~\cite{R_plus_S_g_2P} using different learning rates \cite{deCote:2006} similar to WoLF~\cite{WOLF2002} as well as using embedded emotion~\cite{Emotional_Multiagent}, social networks~\cite{Ohtsuki2006,Santos:2006}.

However, these approaches do not directly apply to Deep RL agents~\cite{Leibo:2017}.
Recent work in this direction~\cite{kleiman2016coordinate,Julien:2017,consequentialist18} focuses on letting agents learn strategies in multi-agent settings through interactions with other agents.  
\citet{Leibo:2017} define the problem of social dilemmas in the Deep RL framework and analyze the outcomes of a fruit-gathering game~\cite{Julien:2017}.
They vary the abundance of resources and the cost of conflict in the fruit environment to generate degrees of cooperation between agents. 
\citet{Hughes:2018} define an intrinsic reward (inequality aversion) that attempts to reduce the difference in obtained rewards between agents. 
The agents are designed to have an aversion to both advantageous (guilt) and disadvantageous (unfairness) reward allocation. 
This handcrafting of loss with mutual fairness evolves cooperation, but it leaves the agent vulnerable to exploitation.
LOLA~\cite{foerster2018learning} uses opponent awareness to achieve high levels of cooperation in the Coin Game and the Iterated Prisoner's Dilemma game.
However, the LOLA agent assumes access to the other agent's policy parameters and gradients. 
This level of access is analogous to getting complete access to the other agent's private information and therefore devising a strategy with full knowledge of how they are going to play. 
\citet{Wang:2019} propose an evolutionary Deep RL setup to evolve cooperation. 
They define an intrinsic reward that is based on features generated from the agent's past and future rewards, and this reward is shared with other agents. They use evolution to maximize the sum of rewards among the agents and thus evolve cooperative behavior. 
However, sharing rewards in this indirect way enforces cooperation rather than evolving it through independently learning agents.

In contrast, we introduce a Status-Quo ($SQLoss$) that evolves cooperation between agents without sharing rewards, gradients, or using a communication channel. 
The $SQLoss$ encourages an agent to imagine the consequences of sticking to the status-quo.
This imagined stickiness ensures that an agent gets a better estimate of a cooperative or selfish policy.
Without $SQLoss$, agents repeatedly switch policies (from cooperative to selfish), obtain short-term rewards (through exploitation), and, therefore incorrectly learn that a selfish strategy gives higher rewards in the long-term.  
\todopb{Figure~??? shows how $SQLoss$ compares to existing approaches to evolve cooperative behavior in multi-agent reinforcement learning games.} 

\begin{figure*}
    \centering
    \includegraphics[width=\linewidth]{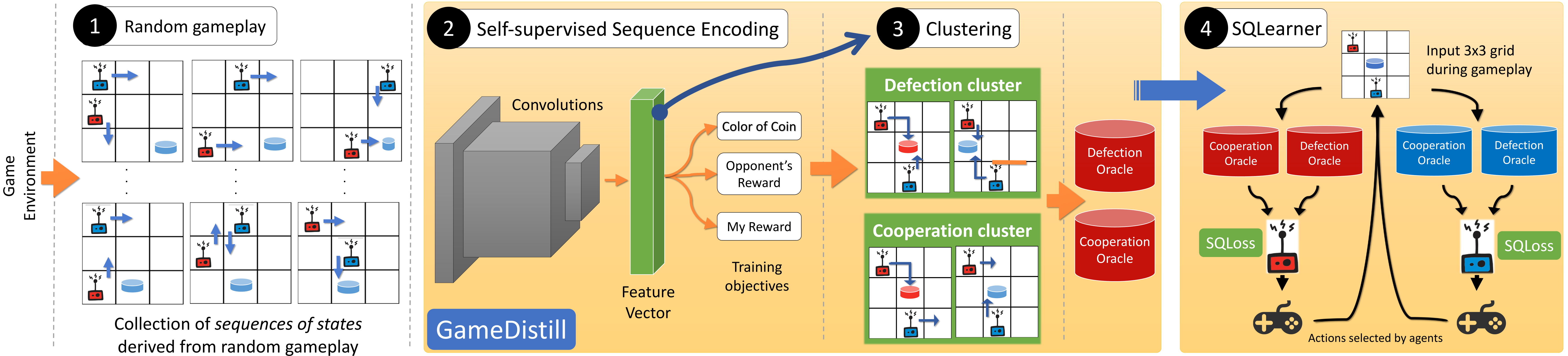}
    \caption{
    High-level architecture of our approach.
    Each agent runs \ipdistill by performing $(1), (2), (3)$ individually before playing the social dilemma game. This creates two oracles per agent.
    During game-play$(4)$, each agent (enhanced with $SQLoss$) takes either the action suggested by the cooperation oracle or the action suggested by the defection oracle. 
    }
    \label{fig:high_level_approach}
\end{figure*}

To work with social dilemma games that have visual observations, we introduce \ipdistill.
\ipdistill uses self-supervision and clustering to automatically extract a cooperative and a selfish policy from a social dilemma game.
The input to \ipdistill is a collection of state sequences derived from game-play between two randomly initialized agents. 
Each state sequence represents a collection of states and actions (of both agents), leading up to a reward in the environment.  
\ipdistill uses this collection of state sequences to learn two oracles.
One oracle represents a cooperative policy, and the other oracle represents a selfish policy.
Given a state, an oracle returns an action according to the specific policy.  
It is important to note that each agent independently runs \ipdistill to extract oracles. 
(For instance, Figure~\ref{fig:oracle_predictions} (Appendix~\ref{appendix:oracle:illustration}) illustrates the cooperation and defection oracles extracted by the Red agent using \ipdistill in the Coin Game.)
Figure~\ref{fig:high_level_approach} shows the high-level architecture of our approach.

\begin{enumerate}
    \item For a social dilemma game with visual observations, each RL agent runs \ipdistill to learn oracles that implement cooperative and selfish behavior. 
    \item We train agents (with $SQLoss$) to play the game such that at any step, an agent can either take the action suggested by the cooperation oracle or the selfish oracle. 
\end{enumerate}

We empirically demonstrate in Section~\ref{sec:results} that our approach evolves cooperative behavior between independently trained agents.

\section{Approach}
\label{sec:approach}

%\begin{figure*}
%    \includegraphics[width=\linewidth]{images/skill_detection_network.png}
%    \caption{Network architecture diagram for skill discovery network}
%    \label{fig:skill_detection_network}
%\end{figure*}

%\begin{figure}
%    \includegraphics[width=\linewidth]{images/action_prediction_network.png}
%    \caption{Neural Network architecture for the Action Prediction Network}
%    \label{fig:action_prediction_network}
%\end{figure}

\subsection{Social Dilemmas modeled as Iterated Matrix Games}
We adopt the definitions in \citet{foerster2018learning}.
We model social dilemmas as general-sum Markov (simultaneous move) games.
A multi-agent Markov game is specified by $G = \langle$$S$, $A$, $U$, $P$, $r$, $n$, $\gamma$$\rangle$.
$S$ denotes the state space of the game. 
$n$ denotes the number of agents playing the game.
At each step of the game, each agent $a \in A$, selects an action $u^a \in U$.
$\Vec{u}$ denotes the joint action vector that represents the simultaneous actions of all agents. 
The joint action $\Vec{u}$ changes the state of the game from $s$ to $s'$ according to the state transition function $P(s'|\Vec{u},s): S \times \textbf{U} \times S \rightarrow  [0, 1]$.
At the end of each step, each agent $a$ gets a reward according to the reward function $r^{a}(s, \Vec{u}): S \times \textbf{U} \rightarrow \mathbb{R}$. 
The reward obtained by an agent at each step is a function of the actions played by all agents. 
For an agent, $a$, the discounted future return from time $t$ is defined as $R_{t}^{a} = \sum_{l=0}^{\infty} \gamma^{l} r_{t+l}^{a}$,
where $\gamma \in [0, 1)$ is the discount factor.
Each agent independently attempts to maximize its expected total discounted reward. 

Matrix games are the special case of two-player perfectly observable Markov games~\cite{foerster2018learning}. 
Table~\ref{tab:payoff_matrix} shows examples of matrix games that represent social dilemmas.
Consider the Prisoner's Dilemma matrix game in Table~\ref{tab:payoff_matrix_ipd}.
Each agent can either cooperate ($C$) or defect ($D$).
For an agent, playing $D$ is the rational choice, regardless of whether the other agent plays either $C$ or $D$.
Therefore, if both agents play rationally, they each receive a reward of $-2$. 
However, if each agent plays $C$, then it will obtain a reward of $-1$. 
This fact that individually rational behavior leads to a sub-optimal group (and individual) outcome highlights the dilemma. 

In Infinitely Iterated Matrix Games, agents repeatedly play a particular matrix game against each other. 
In each iteration of the game, each agent has access to the actions played by both agents in the previous iteration.
Therefore, the state input to an RL agent consists of the actions of both agents in the previous iteration of the game.
We adopt this state formulation to remain consistent with \citet{foerster2018learning}. 
The infinitely iterated variations of the matrix games in Table~\ref{tab:payoff_matrix} represent sequential social dilemmas. 
For ease of representation, we refer to infinitely iterated matrix games as iterated matrix games in subsequent sections. 

\subsection{Learning Policies in Iterated Matrix Games: The Selfish Learner}
\label{sec:approach:selfish-learner}

The standard method to model agents in iterated matrix games is to model each agent as a Deep RL agent that independently attempts to maximize its expected total discounted reward.
Several approaches to model agents in this way use policy gradient-based methods~\cite{sutton2000policy, williams1992simple}).
Policy gradient methods update an agent's policy, parameterized by $\theta^{a}$, by performing gradient ascent on the expected total discounted reward $\E [R_{0}^{a}]$. 
Formally, let $\theta^{a}$ denote the parameterized version of an agent's policy $\pi^{a}$ and $V_{\theta^{1},\theta^{2}}^{a}$ denote the total expected discounted reward for agent $a$. 
Here, $V^{a}$ is a function of the policy parameters $(\theta^{1}, \theta^{2})$ of both agents. 
In the $i^{th}$ iteration of the game, each agent updates $\theta_{i}^{a}$ to $\theta_{i+1}^{a}$, such that it maximizes it's total expected discounted reward. $\theta_{i+1}^{a}$ is computed as follows:
% \vspace{-6pt}
\begin{equation}
    \begin{split}
        \theta_{i+1}^{1} & = argmax_{\theta^{1}} V^{1}(\theta^{1}, \theta_{i}^{2})\\
        \theta_{i+1}^{2} & = argmax_{\theta^{2}} V^{2}(\theta_{i}^{1}, \theta^{2})\\
    \end{split}
\end{equation}
For agents trained using reinforcement learning, the gradient ascent rule to update $\theta_{i+1}^{1}$ is:
\begin{equation}
    \begin{split}
        f_{nl}^{1} & = \nabla_{\theta^{i}_{1}} V^{1} (\theta_{i}^{1}, \theta_{i}^{2}) \cdot \delta \\
        \theta_{i+1}^{1} & = \theta_{i}^{1} + f_{nl}^{1} (\theta_{i}^{1}, \theta_{i}^{2}) 
    \end{split}
    \label{eq:naive_learner_update_rule}
\end{equation}

where $\delta$ is the step size of the updates.
\todopb{they use policy-gradient and actor-critic -- add this information as well, now it does not have much information}

In the Iterated Prisoner's Dilemma (IPD) game, agents trained with the policy gradient update method converge to a sub-optimal mutual defection equilibrium (Figure~\ref{fig:results_IPD}, \cite{Peysakhovich1707.01068}).
This sub-optimal equilibrium attained by Selfish Learners motivates us to explore alternative methods that could lead to a desirable cooperative equilibrium.
We denote the agent trained using policy gradient updates as a Selfish Learner ($SL$).

\begin{figure}
    \centering
    \includegraphics[width=0.85\linewidth]{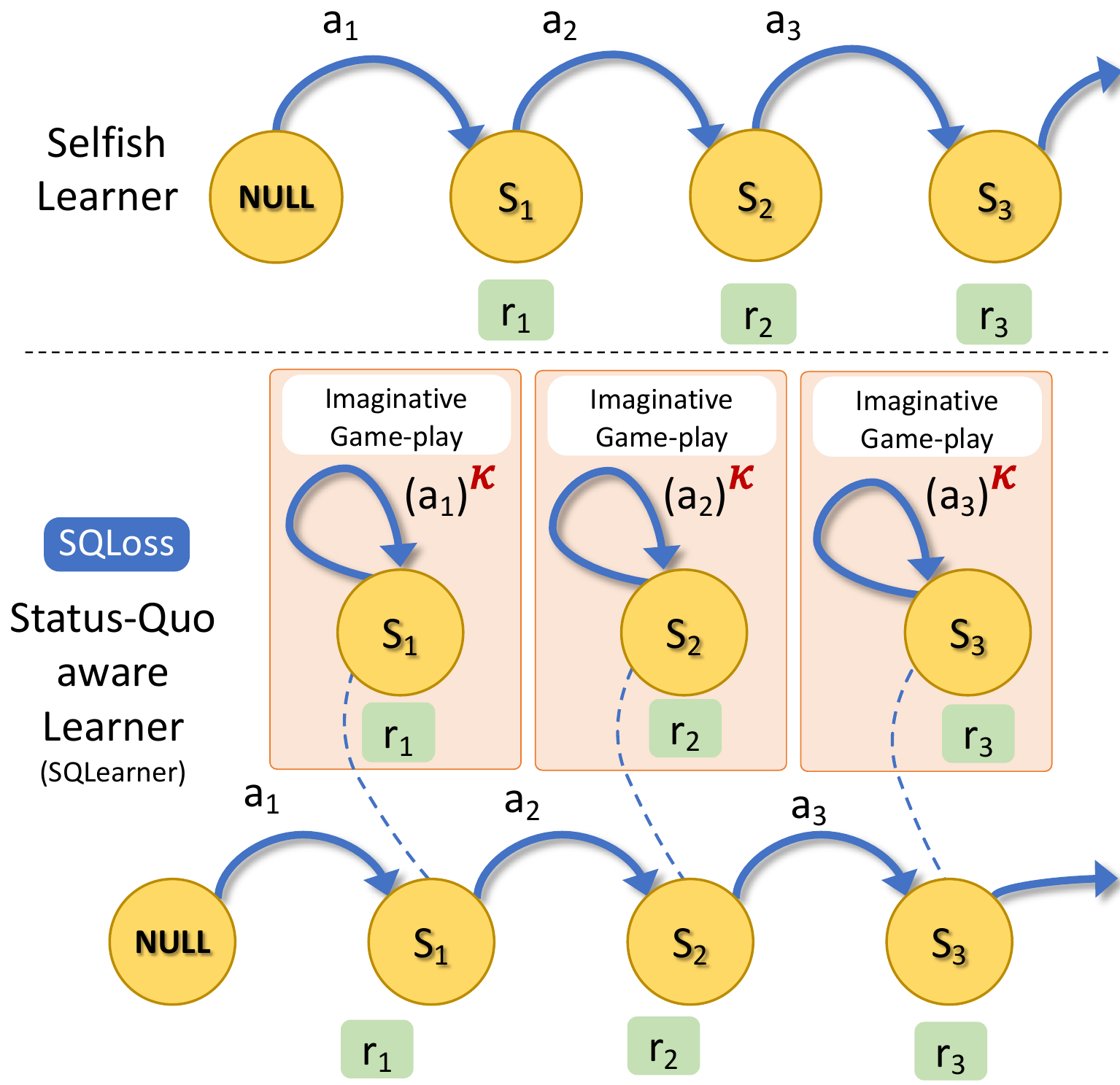}
    \caption{
    Intuition behind $SQLearner$.
    At each step, the $SQLoss$ encourages an agent to imagine the consequences of sticking to the status-quo by imagining an episode where the status-quo is repeated for $\kappa$ steps.  
    Section~\ref{sec:approach:SQLoss} describes $SQLoss$ in more detail. 
    }
    \label{fig:rl_imaginative_learning}
\end{figure}

\subsection{Learning Policies in Iterated Matrix Games: The Status-Quo Aware Learner ($SQLoss$)}
\label{sec:approach:SQLoss}

\subsubsection{$SQLoss$: Intuition}
Why do independent, selfish learners converge to mutually harmful behavior in the IPD?
To understand this, consider the payoff matrix for a single iteration of the IPD in Table~\ref{tab:payoff_matrix_ipd}.
In each iteration, an agent can play either $C$ or $D$.
Mutual defection $(D, D)$ is worse for each agent than mutual cooperation $(C, C)$.
However, one-sided exploitation $(D, C)$ is better than mutual cooperation for the exploiter and far worse for the exploited. 
Therefore, as long as an agent perceives the possibility of exploitation ($(D, C)$ or $(C, D)$), it is drawn to defect, both to maximize reward (through exploitation) and minimize loss (through being exploited). 
To increase the likelihood of cooperation, it is important to reduce instances of exploitation between agents. 
We posit that, if agents mostly only either mutually cooperate $(C, C)$ or mutually defect $(D, D)$, then they will learn to prefer $C$ and achieve a socially desirable cooperative equilibrium.   

Motivated by this idea, we introduce a status-quo loss ($SQLoss$) for each agent derived from the idea of imaginary game-play, as depicted in Figure~\ref{fig:rl_imaginative_learning}. 
Intuitively, the loss encourages an agent to imagine an episode where the status-quo (current situation) is repeated for a number of steps. 
This imagined episode causes the exploited agent (in $(D, C)$) to perceive a continued risk of exploitation and, therefore, quickly move to $(D, D)$.
Hence, for an agent, the short-term gain from exploitation $(D, C)$ is overcome by the long-term loss from mutual exploitation $(D, D)$. 
Therefore, agents move towards either mutual cooperation $(C, C)$ or mutual defection $(D, D)$.
With exploitation (and subsequently, the fear of being exploited) out of the picture, agents move towards mutual cooperation. 
Figure~\ref{fig:rl_imaginative_learning} shows the idea behind $SQLoss$. 

\subsubsection{$SQLoss$: Formulation}
We describe below the formulation of SQLoss with respect to agent 1. The formulation for agent 2 is identical to that of agent 1.
Let $\uptau_a = (s_{0}, u_{0}^{1}, u_{0}^{2}, r_{0}^{1}, \cdots s_{T}, u_{T}^{1}, u_{T}^{2}, r_{T}^{1})$ denote the collection of an agent's experiences after $T$ time steps.
Let $R_{t}^{1} (\uptau_1)$ denote the discounted future return for agent $1$ starting at $s_t$ in actual game-play.
Let $\hat \uptau_{1}$ denote denote the collection of an agent's \textbf{imagined} experiences. 
For a state $s_{t}$ ($t \in [0, T)$), an agent imagines an episode by starting at $s_{t}$ and repeating $u_{t-1}^{1}, u_{t-1}^{2}$ for $\kappa_{t}$ steps.
This is equivalent to imagining a $\kappa_{t}$ step repetition of already played actions. 
We sample $\kappa_{t}$ from a Discrete Uniform distribution $\mathbb{U} \{1,z\}$ where $z$ is a hyper-parameter $\geq 1$.
To simplify notation, let $\phi_t(s_t, \kappa_{t})$ denote the ordered set of state, actions, and rewards starting at time $t$ and repeated $\kappa_{t}$ times for imagined game-play.
Let $\hat R_{t}^{1} (\hat \uptau_1)$ denote the discounted future return starting at $s_t$ in imagined status-quo game-play.
% \vspace{-8pt}
{
\begin{multline}%
    \tiny
    \phi_t(s_t,\kappa_{t}) = \big[ (s_{t}, u_{t-1}^{1}, u_{t-1}^{2}, r_{t-1}^{1})_0, \cdots, \\
    (s_{t}, u_{t-1}^{1}, u_{t-1}^{2}, r_{t-1}^{1})_{\kappa_{t}-1} \big]
    % \label{eq:tau_hat_phi}
\end{multline}
% \vspace{-8pt}
\begin{multline}
    \tiny
    \hat \uptau_{1} = \big( \phi_t(s_t,\kappa_{t}), (s_{t+1}, u_{t+1}^{1}, u_{t+1}^{2}, r_{t+1}^{1})_{\kappa_{t} + 1}, \cdots, \\
    (s_{T}, u_{T}^{1}, u_{T}^{2}, r_{T}^{1})_{T+\kappa_{t}-t} \big)
    % \label{eq:tau_hat}
\end{multline}%

}

% \vspace{-8pt}
{
\begin{equation}
    \small
    \begin{split}
        R_{t}^{1} (\uptau_1) & = \sum_{l=t}^{T} \gamma^{l-t} r_{l}^{1} \\
        \hat R_{t}^{1} (\hat \uptau_1) & = \Big( \frac{1 - \gamma^{\kappa}}{1 - \gamma} \Big) r_{t-1}^{1} + \gamma^{\kappa} R_{t}^{1} (\uptau_1) \\
        & = \Big( \frac{1 - \gamma^{\kappa}}{1 - \gamma} \Big) r_{t-1}^{1} + \gamma^{\kappa} \sum_{l=t}^{T} \gamma^{l-t} r_{l}^{1}
    \end{split}
\end{equation}
}

$V_{\theta^{1},\theta^{2}}^{1}$ and $\hat V_{\theta^{1},\theta^{2}}^{1}$ are approximated by $\E [R_{0}^{1} (\uptau_1)]$ and $\E[\hat R_{0}^{1} (\hat \uptau_1)]$ respectively. 
These $V$ values are the expected rewards conditioned on both agents' policies ($\pi^{1}$, $\pi^{2}$). 
For agent 1, the regular gradients and the Status-Quo gradients,
$\nabla_{\theta^{1}} \E [R_{0}^{1}(\uptau_1)]$ and $\nabla_{\theta^{1}} \E [\hat R_{0}^{1}(\hat \uptau_1)]$, can be derived from the policy gradient formulation as:
\begin{equation}
    % \tiny
    \begin{split}
        \nabla_{\theta^{1}} \E [R_{0}^{1}(\uptau_1)] = \E & [R_{0}^{1}(\uptau_1)\nabla_{\theta^{1}} log\pi^{1}(\uptau_1)] \\
        = \E {\Big[} & \sum_{t=0}^{T} \nabla_{\theta^{1}} log\pi^{1} (u_{t}^{1}|s_{t}) \cdot \sum_{l=t}^{T} \gamma^{l} r_{l}^{1} {\Big]} \\
        = \E {\Big[} & \sum_{t=0}^{T} \nabla_{\theta^{1}} log\pi^{1} (u_{t}^{1}|s_{t}) \gamma^{t} \big( R_{t}^{1}(\uptau_1) - b(s_{t}) \big) \Big] \end{split}
\end{equation}%
\begin{equation}%
    % \tiny
    \begin{split}%
        \nabla_{\theta^{1}} \E [\hat R_{0}^{1}(\hat \uptau_1)] & = \E~[\hat R_{0}^{1}(\hat \uptau_1)\nabla_{\theta^{1}} log\pi^{1}(\hat \uptau_1)] \\
% \cdot
        = \E {\Bigg[} \sum_{t=0}^{T} & \nabla_{\theta^{1}} log\pi^{1} (u_{t-1}^{1}|s_{t}) \times \\
        & ~~~~~~~~~~~~~~~~~~~~~\Bigg( \sum_{l=t}^{t+\kappa} \gamma^{l} r_{t-1}^{1} + \sum_{l=t}^{T} \gamma^{l+\kappa} r_{l}^{1} \Bigg) \Bigg] \\
        = \E {\Bigg[} \sum_{t=0}^{T} & \nabla_{\theta^{1}} log\pi^{1} (u_{t-1}^{1}|s_{t}) \times \\
        & ~~~~~~~~~\Bigg( \Big( \frac{1 - \gamma^{\kappa}}{1 - \gamma} \Big) \gamma^{t} r_{t-1}^{1} + \gamma^{\kappa} \sum_{l=t}^{T} \gamma^{l} r_{l}^{1} \Bigg) \Bigg] \\
        = \E {\Big[} \sum_{t=0}^{T} & \nabla_{\theta^{1}} log\pi^{1} (u_{t-1}^{1}|s_{t}) \gamma^{t} \big( \hat R_{t}^{1}(\hat \uptau_1) - b(s_{t}) \big) \Big]
    \end{split}
    \label{eq:gradient_equation}
\end{equation}

where $b(s_{t})$ is a baseline for variance reduction.

Then the update rule $f_{sql,pg}$ for the policy gradient-based Status-Quo Learner (SQL-PG) is,
% \vspace{-4pt}
\begin{equation}
    \small
    f_{sql,pg}^{1} = \big( \alpha \cdot \nabla_{\theta^{1}} \E [R_{0}^{1}(\uptau_1)] + \beta \cdot \nabla_{\theta^{1}} \E [\hat R_{0}^{1}(\uptau_1)] \big) \cdot \delta
\end{equation}
% \vspace{-2pt}
where $\alpha$ and $\beta$ denote the loss scaling factor for the reinforce and the imaginative game-play, respectively.

\subsection{\ipdistill: Moving Beyond Iterated Matrix Games}
\label{sec:GameDistill}
In the previous sections, we have focused on evolving cooperative behavior in the iterated matrix game formulation of sequential social dilemmas. 
In the iterated matrix game formulation, an agent is only allowed to either cooperate or defect in each iteration.
However, in a social dilemma game with visual observations, it is not clear what set of low-level actions constitute cooperative or selfish behavior. 
Therefore, to work on social dilemmas with visual observations, we propose \ipdistill, an approach that automatically learns a cooperation and a defection policy by analyzing the behavior of randomly initialized agents.  
\ipdistill learns these policies in the form of cooperation and defection oracles. 
Given a state, the cooperation oracle suggests an action that represents cooperative behavior. 
Similarly, the defection oracle suggests an action that represents selfish behavior (Figure~\ref{fig:oracle_predictions} (Appendix~\ref{appendix:oracle:illustration})). 
When RL agents play the social dilemma game, each agent independently runs \ipdistill before playing the game. 
Once both agents have run \ipdistill, they consult either of the two extracted oracles in every step of the game. 
Therefore, in each step, an agent either takes the action recommended by the cooperation oracle or the action recommended by the defection oracle.
In this way, we reduce the visual input game to an iterated matrix game and subsequently apply $SQLoss$ to evolve cooperative behavior. 
\ipdistill (see Figure~\ref{fig:high_level_approach}) works as follows. 
\begin{enumerate}
    \item 
    We initialize RL agents with random weights and play them against each other in the game. 
    In these random game-play episodes, whenever an agent receives a reward, we store the sequence of the last three states up to the current state. 
    \item
    This collection of state sequences is used to train the \ipdistill network. 
    The \ipdistill network takes as input a sequence of states and predicts the rewards of both agents as well as environment parameters that depend on the game. 
    For instance, in the Coin Game, the \ipdistill network predicts the rewards of both agents and the color of the picked coin.
    \item 
    Training the \ipdistill network leads to the emergence of feature embeddings for the various state sequences. Subsequently, clustering these embeddings using Agglomerative Clustering (with number of clusters=2) leads to the development of cooperative and defection clusters. One of the learned clusters contains state sequences that represent cooperative behavior, and the other cluster contains state sequences that represent defection.
    For instance, in the Coin Game, a point in the cooperation cluster contains a sequence of states where an agent picks a coin of its color. 
    \item
    To train the cooperation and defection oracle networks, we use the collection of state sequences in each cluster. 
    For each sequence of states in a cluster, we train the oracle network to predict the next action, given the current state. 
    For instance, Figure~\ref{fig:oracle_predictions} (Appendix~\ref{appendix:oracle:illustration})shows the cooperation and defection oracles extracted by the Red agent using \ipdistill in the Coin Game.
     
\end{enumerate}

Section~\ref{sec:experimental-setup:gamedistill} describes the architectural choices of each component of \ipdistill. 
 
\begin{comment}
For any game, the idea is summarized as follows:
\begin{enumerate}
    \item \textbf{N complex actions -> M complex skills:} Learning decoupled skills has multiple benefits as they can be easily transferred to other agents, or new skills can be learned, or skills can be updated.  This also decouples the environment-specific information from the skills of the agents.
    \item \textbf{Action Prediction Network:} Once the skills are identified, this module converts the meta-learning (in the form of skill) to environment-specific actions/interactions. This allows the agents to behave a certain way in the environment given a skill.
    \item \textbf{Skill selection policy:} Given agents can interact with the environment given a skill/policy, now the task boils down to selecting a skill at a given state that maximizes the overall reward of the agents.
\end{enumerate}
\end{comment}

\begin{comment}

\begin{figure}[t]
    \centering
    \includegraphics[width=0.4\textwidth]{images/meta_policy.png}
    \caption{Reduction of a game to an IPD formulation. We summarize complex skills using oracles and the agent has to learn a meta-policy to switch between skills. \todopb{Make a formal version of this reduction}}
    \label{fig:meta_policy}
\end{figure}

\end{comment}

\section{Experimental Setup}
\label{sec:experimental_setup}

\begin{table*}
    \centering
    \small
    \begin{subtable}[t]{0.3\linewidth}
        \centering
        \renewcommand{\arraystretch}{1.2}
        \begin{tabular}{c|c|c}
             & $C$ & $D$ \\ \hline
             $C$ & (-1, -1) & (-3, 0) \\ \hline
             $D$ & (0, -3) & (-2, -2) \\ \hline
        \end{tabular}
        \caption{Prisoners' Dilemma (PD)}
        \label{tab:payoff_matrix_ipd}
    \end{subtable}%
    \begin{subtable}[t]{0.3\linewidth}
        \centering
        \renewcommand{\arraystretch}{1.2}
        \begin{tabular}{c|c|c}
             & $H$ & $T$ \\ \hline
             $H$ & (+1, -1) & (-1, +1) \\ \hline
             $T$ & (-1, +1) & (+1, -1) \\ \hline
        \end{tabular}
        \caption{Matching Pennies (MP)}
        \label{tab:payoff_matrix_imp}
    \end{subtable}%
    \begin{subtable}[t]{0.3\linewidth}
        \centering
        \renewcommand{\arraystretch}{1.2}
        \begin{tabular}{c|c|c}
             & $C$ & $D$ \\ \hline
             $C$ & (0, 0) & (-4, -1) \\ \hline
            %  $C$ & (4, 4) & (0, 3) \\ \hline
             $D$ & (-1, -4) & (-3, -3) \\ \hline
            %  $D$ & (3, 0) & (1, 1) \\ \hline
        \end{tabular}
        \caption{Stag Hunt (SH)}
        \label{tab:payoff_matrix_sh}
    \end{subtable}%
    
    \caption{\label{tab:payoff_matrix}%
    Payoff matrices for the different games used in our experiments. 
    $(X,Y)$ in a cell represents a reward of $X$ to the row and $Y$ to the column player. 
    $C$, $D$, $H$, and $T$ denote the actions for the row and column players.
    In the iterated versions of these games, agents play against each other over several iterations. 
    In each iteration, an agent takes an action and receives a reward based on the actions of both agents. 
    Each matrix represents a different kind of social dilemma.
    }
\end{table*}

In order to compare our results to previous work~\cite{foerster2018learning}, we use the Normalized Discounted Reward ($NDR = (1-\gamma) \sum_{t=0}^{T} \gamma^{t} r_{t}$). 
A higher NDR implies that an agent obtains a higher reward in the environment. 
We compare our approach (Status-Quo Aware Learner: $SQLearner$) to Learning with Opponent-Learning Awareness (Lola-PG)~\cite{foerster2018learning} and the Selfish Learner (SL, Section~\ref{sec:approach:selfish-learner}).
For all experiments, we perform $20$ runs and report average $NDR$, along with variance across runs.
The bold line in all the figures is the mean, and the shaded region is the one standard deviation region over the mean. 
All of our code is available at~\citet{github_code}.

\subsection{Social Dilemma Games}
For our experiments with social dilemma matrix games, we use the (Iterated Prisoners Dilemma (IPD)~\cite{luce1989games}, Iterated Matching Pennies (IMP)~\cite{lee1967application}, and the Iterated Stag Hunt (ISH)~\cite{foerster2018learning}). 
Table~\ref{tab:payoff_matrix} shows the payoff matrix for a single iteration of each game. 
In iterated matrix games, at each iteration, agents take an action according to a policy and receive the rewards in Table~\ref{tab:payoff_matrix}.
To simulate an infinitely iterated game, we let agents play 200 iterations of the game against each other, and do not provide an agent with any information about the number of remaining iterations~\cite{foerster2018learning}. 
In an iteration, the state for an agent is the actions played by both agents in the previous iteration. 
Each matrix game in Table~\ref{tab:payoff_matrix} represents a different dilemma. 

In the Prisoner's Dilemma, the rational policy for each agent is to defect, regardless of the policy of the other agent.  
However, when each agent plays rationally, each is worse off. 
In Matching Pennies, if an agent plays predictably, it is prone to exploitation by the other agent.
Therefore, the optimal policy is to randomize between $H$ and $T$, obtaining an average NDR of $0$.
The Stag Hunt game represents a coordination dilemma. 
In the game, given that the other agent will cooperate, an agent's optimal action is to cooperate as well. 
However, at each step, each agent has an attractive alternative, that of defecting and obtaining a guaranteed reward of $3$. 
Therefore, the promise of a safer alternative and the fear that the other agent might select the safer choice could drive an agent to also select the safer alternative, thereby sacrificing the higher reward of mutual cooperation. 

For our experiments on a social dilemma with visual observations, we use the Coin Game (Figure~\ref{fig:coin_game}) \cite{foerster2018learning}.
The rational policy for an agent is to defect and try to pick all coins, regardless of the policy of the other agent.
However, when both agents defect, both are worse off. 

\subsection{SQLoss}
For our experiments with the Selfish and Status-Quo Aware Learner ($SQLearner$), we use policy gradient-based learning where we train an agent with the Actor-Critic method~\cite{sutton2011reinforcement}. 
Each agent is parameterized with a policy actor and critic for variance reduction in policy updates. 
During training, we use gradient descent with step size, $\delta = 0.005$ for the actor and $\delta = 1$ for the critic. 
We use a batch size of $4000$ for Lola-PG~\cite{foerster2018learning} and 200 for $SQLearner$ for roll-outs. 
We use an episode length of 200 for all iterated matrix games. 
We use a discount rate ($\gamma$) of $0.96$ for the Iterated Prisoners' Dilemma, Iterated Stag Hunt, and Coin Game. 
For the Iterated Matching Pennies, we use $\gamma = 0.9$.
The high value of $\gamma$ allows for long time horizons, thereby incentivizing long-term reward. 
Each agent randomly samples $\kappa$ from $\mathbb{U} \in (1, z)$ ($z=10$, discussed in Appendix~\ref{appendix:effect-z-convergence}) at each step.

\subsection{GameDistill}
\label{sec:experimental-setup:gamedistill}

\ipdistill consists of two components.
First, the state sequence encoder (Step 2, Section~\ref{sec:GameDistill}) that takes as input a sequence of states and outputs a feature representation. 
We encode each state in the sequence using a series of standard Convolution layers with kernel-size 3. 
We then use a fully-connected layer with 100 neurons that outputs a dense representation of the sequence of states.
The color of picked coin, agent reward, and opponent reward branches consist of a series of dense layers with linear activation.
We use linear activation so that we can cluster the feature vectors (embeddings) using a linear clustering algorithm, such as Agglomerative Clustering. 
We obtain similar results when we use the K-means clustering algorithm. 
We use the \textit{Binary-Cross-Entropy (BCE)} loss for classification and the \textit{mean-squared error (MSE)} loss for regression. 
We use the \textit{Adam}~\cite{kingma2014adam} optimizer (learning rate $0.003$).

Second, the oracle network (Step 4, Section~\ref{sec:GameDistill}), that predicts an action for an input state. 
We encode the input state using $2$ convolution layers with kernel-size $3$ and \textit{relu} activation. 
To predict the action, we use $3$ fully-connected layers with \textit{relu} activation and the BCE loss.
We use \textit{L2} regularization, and \textit{Gradient Descent} with the \textit{Adam} optimizer (learning rate $0.001$).

\section{Results}
\label{sec:results}

\subsection{Learning optimal policies in Iterated Matrix Games using SQLoss}
\label{sec:results:iterated-matrix-games:sqloss}

\paragraph{\textbf{Iterated Prisoner's Dilemma (IPD):}}
\begin{figure}
    \centering
    \includegraphics[width=0.7\linewidth]{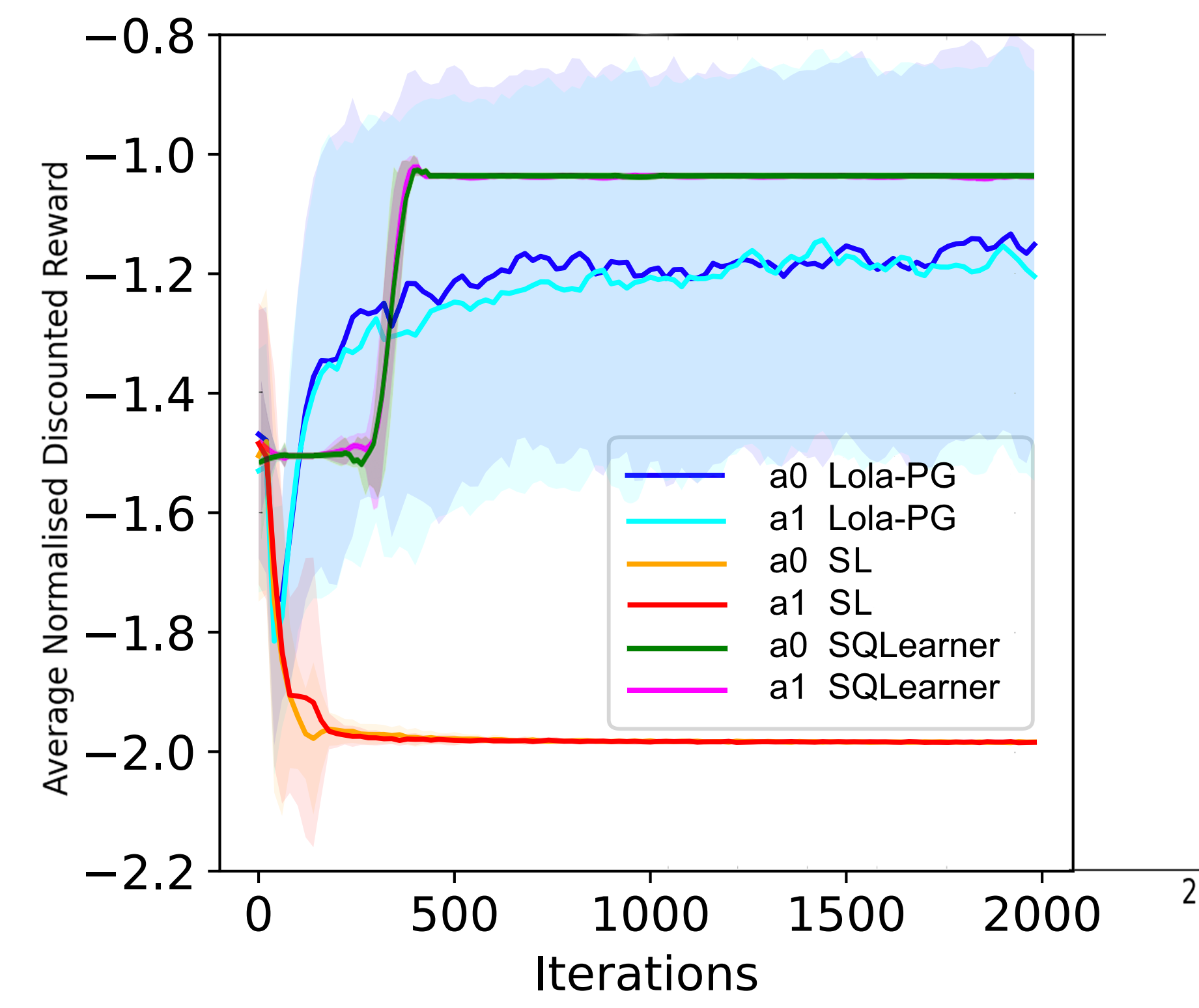}
    \caption{\label{fig:results_IPD}%
    Average NDR values for different learners in the IPD game. 
    $SQLearner$ agents obtain a near-optimal NDR value $(-1)$ for this game. 
    In contrast to other methods, $SQLearner$ agents have close to zero variance across runs. 
    }
\end{figure}

We train different learners to play the IPD game. 
Figure~\ref{fig:results_IPD} shows the results.
For all learners, agents initially defect and move towards an NDR of $-2.0$. 
This initial bias towards defection is expected, since, for agents trained with random game-play episodes, the benefits of exploitation outweigh the costs of mutual defection.
For Selfish Learner (SL) agents, the bias intensifies, and the agents converge to mutually harmful selfish behavior (NDR=$-2.0$).  
Lola-PG agents learn to predict each other's behavior and therefore realize that defection is more likely to lead to mutual harm than selfish benefit.
They subsequently move towards cooperation, but occasionally defect (NDR=$-1.2$).
In contrast, $SQLearner$ agents quickly realize the costs of defection, indicated by the small initial dip in the NDR curves.
They subsequently move towards almost perfect cooperation, with an NDR of $-1.0$.
Finally, it is important to note that $SQLearner$ agents have close to zero variance, unlike other methods where the variance in NDR across runs is significant. 

\paragraph{\textbf{Iterated Matching Pennies (IMP):}}
\begin{figure}
    \centering
    \includegraphics[width=\linewidth]{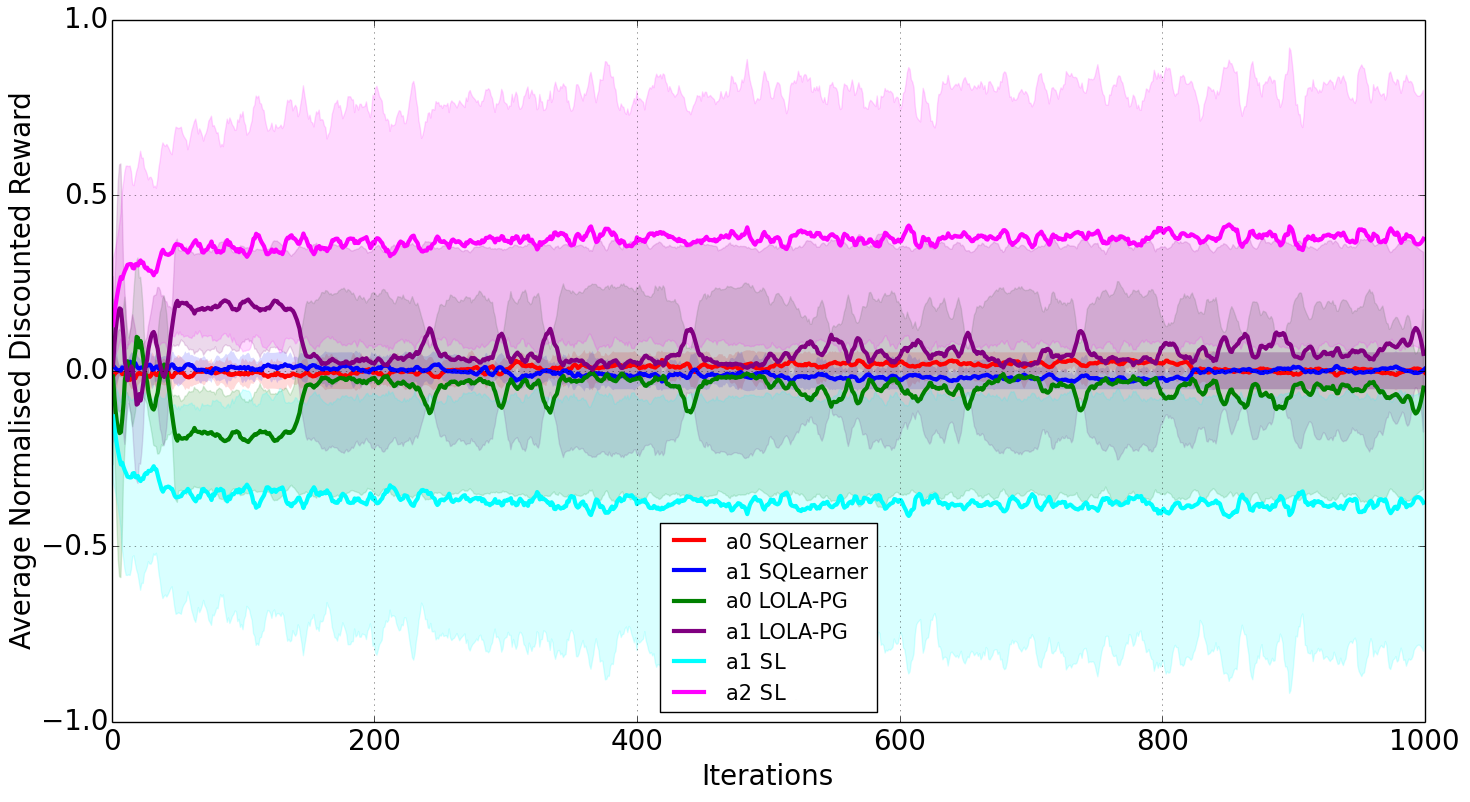}
    \caption{
    Average NDR values for different learners in the IMP game. 
    $SQLearner$ agents avoid exploitation by randomising between $H$ and $T$ to obtain a near-optimal NDR value (0) for this game.
    In contrast to other methods, $SQLearner$ agents have close to zero variance across runs. 
    }
    \label{fig:results_IMP}
\end{figure}

We train different learners to play the IMP game.
The optimal policy for an agent to avoid exploitation is to play $H$ or $T$ perfectly randomly and obtain an NDR of $0$.
Figure~\ref{fig:results_IMP} shows the results. 
$SQLearner$ agents learn to play optimally and obtain an NDR close to $0$. 
Interestingly, Selfish Learner and Lola-PG agents converge to an exploiter-exploited equilibrium where one agent consistently exploits the other agent. 
This asymmetric exploitation equilibrium is more pronounced for Selfish Learner agents than for Lola-PG agents.
As before, we observe that $SQLearner$ agents have close to zero variance across runs, unlike other methods where the variance in NDR across runs is significant. 

Appendix \ref{appendix:results:ish} shows the results for the ISH game. 

\subsection{Evolving Cooperation in Games with visual input using GameDistill followed by SQLoss}
\label{sec:results:visual-input-gamedistll-and-sqloss}

\subsubsection{The Coin Game: GameDistill}
{
\begin{figure}
    \centering
    \includegraphics[width=\linewidth]{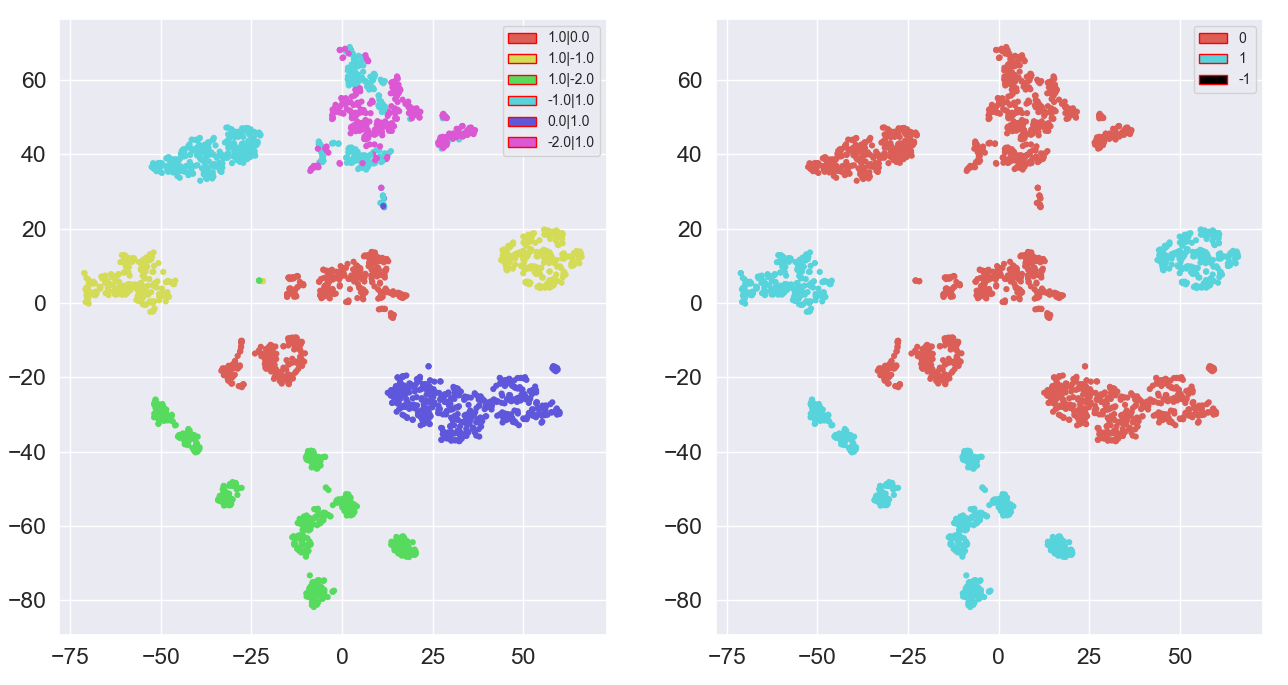}
    \caption{
Representation of the clusters learned by $GameDistill$.
Each point is a t-SNE projection of the 100-dimensional feature vector output by the \ipdistill network for an input sequence of states.  
The figure on the left is colored based on actual rewards obtained by each agent (Red agent followed by Blue agent).
The figure on the right is colored based on clusters learned by $GameDistill$. 
$GameDistill$ correctly identifies two types of state sequences, one for cooperation (blue cluster) and the other for defection (red cluster). 
}
\label{fig:feature_vector_plots}%
\end{figure}
}

\begin{comment}
\begin{figure}[t]
    \centering
    % \begin{subfigure}[t]{0.4\linewidth}
    %     \includegraphics[width=\linewidth]{images/results__CoinGame__IPD_formulation.png}
    %     \caption{Average NDR}
    %     \label{fig:results_CoinGame_IPD_formulation}
    % \end{subfigure}\hfil%
    \begin{subfigure}[t]{\linewidth}
        \includegraphics[width=\linewidth]{images/CoinGame_results_corrected.png}
        \caption{
        Probability of an agent eating a coin of its color
        }
        \label{fig:results_CoinGame_probability_of_eating_coin}
    \end{subfigure}
    \caption{\label{fig:coingame_results}%
    Average NDR (a) and cooperation level (b) for learners trained in the Coin Game. 
    $SQLearner$ agents cooperate (eat only their own coins) to obtain near optimal NDR (-0.5) in the game. 
    }
\end{figure}
\end{comment}

\begin{figure}[t]
    \centering
    \includegraphics[width=0.65\linewidth]{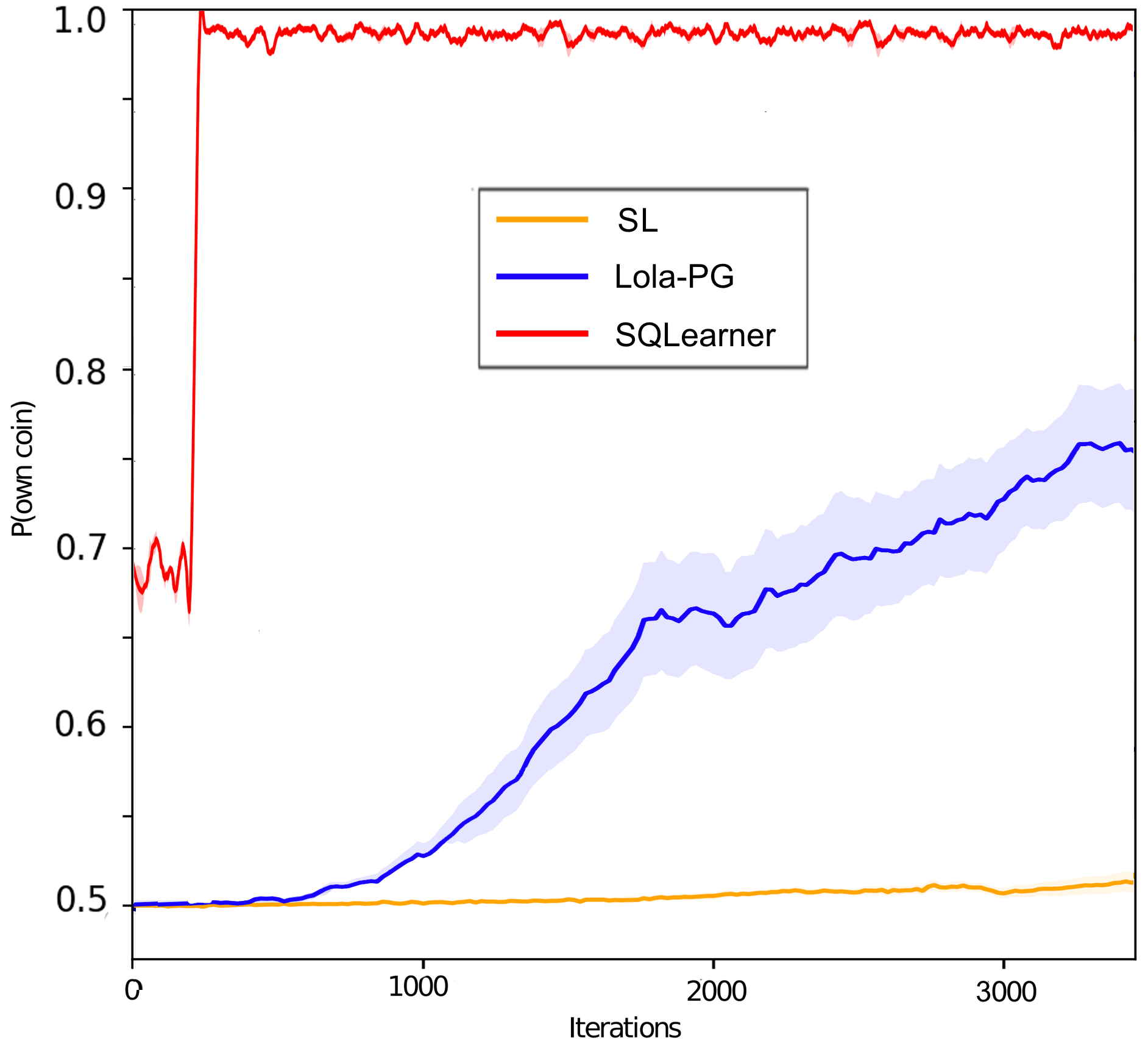}
    \caption{\label{fig:coingame_results}%
    Probability of an agent picking a coin of its color for learners trained in the Coin Game. 
    $SQLearner$ agents cooperate (pick only their own coins) to achieve a near optimal strategy in the game. 
    In contrast to Lola-PG, $SQLearner$ agents have close to zero variance across runs. 
    }
\end{figure}

To evaluate the clustering step in \ipdistill, we make two t-SNE~\cite{maaten2008visualizing} plots of the 100-dimensional feature vector extracted from the last layer of the \ipdistill network. 
The first plot colors each point (state sequence) by the rewards obtained by both agents in the sequence.
The second plot colors each point by the cluster label output by Agglomerative clustering.
Figure~\ref{fig:feature_vector_plots} shows the results. 
\ipdistill correctly learns two clusters, one for state sequences that represent cooperation and the other for state sequences that represent defection.
We also experiment with different values for feature vector dimensions and obtain similar clustering results with sufficient training. 
Once we have the clusters, we train oracle networks using the state sequences in each cluster.
To evaluate that the trained oracles represent a cooperation and a defection policy, we modify the Coin Game environment to contain only the Red agent.
We then play two variations of the game.
In the first variation, the Red agent is forced to play the action suggested by the first oracle. 
In this variation, we find that the Red agent picks only $8.4\%$ of Blue coins, indicating a high rate of cooperation. 
Therefore, the first oracle represents a cooperation policy. 
In the second variation, the Red agent is forced to play the action suggested by the second oracle. 
In this case, we find that the Red agent picks $99.4\%$ of Blue coins, indicating a high rate of defection. 
Hence, the second oracle represents a defection policy.
Therefore, the oracles learned by the Red agent using \ipdistill represent cooperation and defection policies. 

\subsubsection{The Coin Game: SQLoss}
Before playing the game, each $SQLearner$ agent uses \ipdistill to learn cooperation and defection oracles. 
During game-play, at each step, an agent follows either the action suggested by its cooperation oracle or the action suggested by its defection oracle. 
Further, each $SQLearner$ agent has an additional $SQLoss$ term.
We compare approaches using the degree of cooperation between agents, measured by the probability of an agent to pick the coin of its color~\cite{foerster2018learning}.  
Figure~\ref{fig:coingame_results} shows the results. 
The probability that an $SQLearner$ agent will pick the coin of its color is close to $1$. 
This high probability indicates that the other $SQLearner$ agent is cooperating with this agent and only picking coins of its color. 
In contrast, the probability that a Lola-PG agent will pick its own coin is much smaller, indicating higher rates of defection.
As expected, the probability of an agent picking its own coin is the smallest for selfish learners (SL). 
The probability value of $0.5$ indicates that a selfish learner is just as likely to pick the other agent's coin as it is to pick its own coin. 

\subsection{SQLearner: Exploitability and Adaptability}
Given that an agent does not have any prior information about the other agent, it is important that it evolves its strategy based on the strategy of its opponent. 
To evaluate an $SQLearner$ agent's ability to avoid exploitation by a selfish agent, we train one $SQLearner$ agent against an agent that always defects in the Coin Game. 
We find that the $SQLearner$ agent also learns to always defect.
This persistent defection is important since given that the other agent is selfish, the $SQLearner$ agent can do no better than also be selfish. 
To evaluate an $SQLearner$ agent's ability to exploit a cooperative agent, we train one $SQLearner$ agent with an agent that always cooperates in the Coin Game. 
In this case, we find that the $SQLearner$ agent learns to always defect.
This persistent defection is important since given that the other agent is cooperative, the $SQLearner$ agent obtains maximum reward by behaving selfishly. 
Hence, the $SQLearner$ agent is both resistant to exploitation and able to exploit, depending on the strategy of the other agent.

\section{Conclusion}
We have described a status-quo loss ($SQLoss$) that encourages an agent to imagine the consequences of sticking to the status-quo.
We demonstrated how agents trained with $SQLoss$ evolve cooperative behavior in several social dilemmas without sharing rewards, gradients, or using a communication channel. 
To work with visual input games, we proposed $GameDistill$, an approach that automatically extracts a cooperative and a selfish policy from a social dilemma game.
We combined $GameDistill$ and $SQLoss$ to demonstrate how agents evolve desirable cooperative behavior in a social dilemma game with visual observations.
\label{sec:conclusion}

\bibliography{references}
\bibliographystyle{icml2020}

\clearpage

\appendix
\section*{Supplementary Material}

\section{Illustrations of Trained Oracle Networks for the Coin Game}
\label{appendix:oracle:illustration}

\begin{figure}[h]
    \centering
    \includegraphics[width=0.9\linewidth]{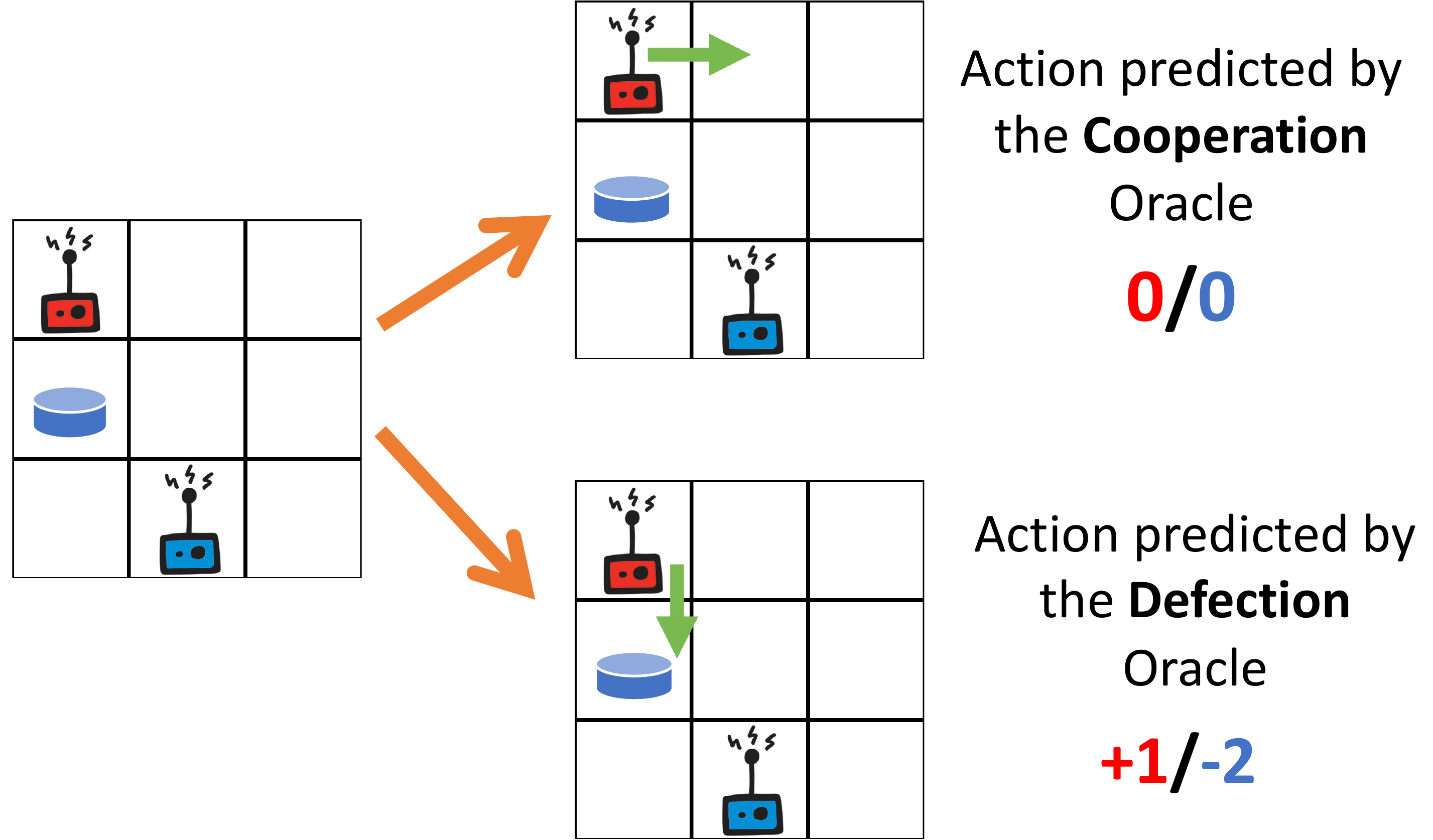}
    \caption{
    Illustrative predictions of the oracle networks learned by the Red agent using \ipdistill in the Coin Game.
    The cooperation oracle suggests an action that avoids picking the coin of the other agent. 
    The defection oracle suggests an action that involves picking the coin of the other agent.
    }
    \label{fig:oracle_predictions}
\end{figure}

Figure~\ref{fig:oracle_predictions} shows the predictions of the oracle networks learned by the Red agent using \ipdistill in the Coin Game.
We see that the cooperation oracle suggests an action that avoids picking the coin of the other agent (the Blue coin). 
Analogously, the defection oracle suggests a selfish action that picks the coin of the other agent.

\section{Results for the Iterated Stag Hunt using SQLoss}
\label{appendix:results:ish}

\begin{figure}[h]
    \centering
    \includegraphics[width=\linewidth]{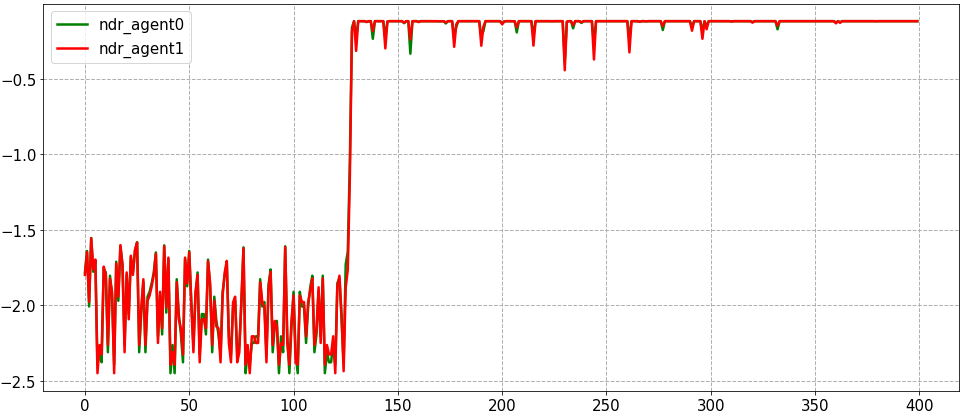}
    \caption{
    NDR values for $SQLearner$ agents in the ISH game.
    $SQLearner$ agents coordinate successfully to obtain a near optimal NDR value ($0$) for this game.
    }
\label{fig:results_stag_hunt}%
\end{figure}

Figure~\ref{fig:results_stag_hunt} shows the results of training two $SQLearner$ agents on the Iterated Stag Hunt game.
$SQLearner$ agents coordinate successfully to obtain a near-optimal NDR value ($0$) for this game.

\section{$SQLoss$: Effect of $z$ on convergence to cooperation}
\label{appendix:effect-z-convergence}

We explore the effect of the hyper-parameter $z$ (Section~\ref{sec:approach}) on convergence to cooperation.
To imagine the consequences of maintaining the status-quo, each agent samples $\kappa_{t}$ from the Discrete Uniform distribution $\mathbb{U} \{1,z\}$. 
Therefore, a larger value of $z$ implies a larger value of $\kappa_{t}$ and longer imaginary episodes. 
We find that larger $z$ (and hence $\kappa$) leads to faster cooperation between agents in the IPD and Coin Game.
This effect plateaus at $z=10$, which we select for our experiments.

\begin{comment}

\begin{figure}[h]
    \centering
    \includegraphics[width=0.8\linewidth]{images/stag_hunt_tsne.png}
    \caption{
    t-SNE plot for stag hunt
    }
\label{fig:stag_hunt_cluster_plot}%
\end{figure}
\end{comment}

\section{Architecture Details}
\label{sec:sup_arch}
We performed all our experiments on an AWS instance with the following specifications.
\begin{itemize}
    \item Model name: Intel(R) Xeon(R) Platinum 8275CL CPU @ 3.00GHz
    \item RAM: 189GB
    \item CPU(s): 96
    \item Architecture: x86\_64
    \item Thread(s) per core: 2
\end{itemize}

\begin{comment}
\begin{algorithm}
\caption{Algorithm for inducing cooperation using $GameDistill$ and $SQLoss$ }
\textbf{Input}: Game environment with multiple agents
\textbf{GameDistill}:

\label{algo:Distill_Sqloss}
% \vspace{-4pt}
\end{algorithm}
\end{comment}
\section{Reproducibility Checklist}
We follow the reproducibility checklist from \href{https://www.cs.mcgill.ca/~jpineau/ReproducibilityChecklist.pdf~}~\cite{reproducibility_checklist} and include further details here. For all the models and algorithms we have included details that we think would be useful for reproducing the results of this work.
\begin{itemize}
\item For all \textbf{models} and \textbf{algorithms} presented, check if you include:
\begin{enumerate}
    \item \textit{A clear description of the mathematical setting, algorithm, and/or model}: \textbf{Yes}. The algorithm is described in detail in Section~\ref{sec:approach}, with all the loss functions used for training being clearly defined. The details of the architecture, hyperparameters used and other algorithm details are given in Section~\ref{sec:experimental_setup}. Environment details are explained in the sections that they are introduced in.
    \item \textit{An analysis of the complexity (time, space, sample size) of any algorithm}: \textbf{No}. We do not include a formal complexity analysis of our algorithm. However, we do highlight the additional computational steps (in terms of losses and parameter updates) in Section~\ref{sec:approach} over standard multi-agent independently learning RL algorithms that would be needed in our approach.
    \item \textit{A link to a downloadable source code, with specification of all dependencies, including external libraries.}: \textbf{Yes}. We have made the source code available at \citet{github_code}.
\end{enumerate}

\item For any \textbf{theoretical claim}, check if you include:
\begin{enumerate}
    \item \textit{A statement of the result}: \textbf{NA}. Our paper is primarily empirical and we do not have any major theoretical claims. Hence this is Not Applicable.
    \item \textit{A clear explanation of any assumptions}: \textbf{NA}.  
    \item \textit{A complete proof of the claim}: \textbf{NA}.
\end{enumerate}
 
\item For all \textbf{figures} and \textbf{tables} that present empirical results, check if you include:
\begin{enumerate}
    \item \textit{A complete description of the data collection process, including sample size}: \textbf{NA}. We did not collect any data for our work.
    \item \textit{A link to a downloadable version of the dataset or simulation environment}: \textbf{Yes}. We have made the source code available at \citet{github_code}.
    \item \textit{An explanation of any data that were excluded, description of any pre-processing step}: \textbf{NA}. We did not perform any pre-processing step.
    \item \textit{An explanation of how samples were allocated for training / validation / testing}: \textbf{Yes}. For \ipdistill the details regarding data used for training is given in Section~\ref{sec:GameDistill}. The number of iterations used for learning (training) by $SQLearner$ is given in Figures~\ref{fig:results_IPD},~\ref{fig:results_IMP} and~\ref{fig:coingame_results}. The details of the number of runs and the batch sizes used for various experiments are given in Section~\ref{sec:experimental_setup}.
    \item \textit{The range of hyper-parameters considered, method to select the best hyper-parameter configuration, and specification of all hyper-parameters used to generate results}: \textbf{Yes}. We did not do any hyperparameter tuning as part of this work. All the hyperparameters that we used are specified in Section~\ref{sec:experimental_setup}.
    \item \textit{The exact number of evaluation runs}: \textbf{Yes}. For all our environments, we repeat the experiment $20$ times. For evaluation of performance, we use an average of $200$ Monte Carlo estimates. We state this in Section~\ref{sec:experimental_setup}. We do not fix any seeds. The details of the number of runs and the batch sizes used for various experiments are also given here.
    \item \textit{A description of how experiments were run}: \textbf{Yes}. The README with instructions on how to run the experiments along with the source code is provided at~\citet{github_code}.
    \item \textit{A clear definition of the specific measure or statistics used to report results}: \textbf{Yes}. We plot the mean and the one standard deviation region over the mean for all our numerical experiments. This is stated in Section~\ref{sec:experimental_setup}.
    \item \textit{Clearly defined error bars}: \textbf{Yes}. We plot the mean and the one standard deviation region over the mean for all our numerical experiments. This is stated in Section~\ref{sec:experimental_setup}.
    \item \textit{A description of results with central tendency (e.g. mean) \& variation (e.g. stddev)}: \textbf{Yes}. We plot the mean and the one standard deviation region over the mean for all our numerical experiments. This is stated in Section~\ref{sec:experimental_setup}.
    \item \textit{A description of the computing infrastructure used}: \textbf{Yes}. We have provided this detail in the Supplementary material in Section~\ref{sec:sup_arch}.
\end{enumerate}
\end{itemize}
\end{document}